\documentclass{sig-alternate-05-2015}

\usepackage{amsmath}
\usepackage{amssymb}
\usepackage{color}
\usepackage{subfigure}
\usepackage{url}
\usepackage{graphicx}
\usepackage{enumitem}
\usepackage[all]{nowidow}
\usepackage{comment}
\usepackage{tikz}
\usetikzlibrary{bayesnet}
\usepackage{algorithm}
\usepackage{algorithmic}
\usepackage{hyphenat}

\graphicspath{{graphics/}}

\usepackage{multirow}

\setlist[enumerate]{itemsep=0.5pt, wide=\parindent}

\def\const{\mathop{\sf const}}

\def\trace{\mathop{\sf tr}}

\def\0{\mathbf{0}}
\def\a{\mathbf{a}}
\def\b{\mathbf{b}}

\def\v{\mathbf{v}}
\def\x{\mathbf{x}}

\def\z{\mathbf{z}}
\def\A{\mathbf{A}}
\def\B{\mathbf{B}}

\def\I{\mathbf{I}}

\def\L{\mathbf{L}}
\def\P{\mathbf{P}}
\def\R{\mathbf{R}}
\def\S{\mathbf{S}}

\def\V{\mathbf{V}}

\def\Z{\mathbf{Z}}

\def\Lambda{\boldsymbol{\lambda}}

\def\bphi{\boldsymbol{\phi}}

\def\Vline{\underline{\V}}

\def\Acal{\mathcal{A}}
\def\Ccal{\mathcal{C}}

\def\Ncal{\mathcal{N}}

\def\Pcal{\mathcal{P}}

\def\Tcal{\mathcal{T}}

\def\Ycal{\mathcal{Y}}

\definecolor{Red}{rgb}{1.0,0.0,0.0}
\newcommand{\todo}[1]{{\color{Red}{\textsf{to do:} \emph{#1}}}}

\let\oldbibliography\thebibliography
\renewcommand{\thebibliography}[1]{%
  \oldbibliography{#1}%
  \setlength{\itemsep}{-0.2pt}%
}

\title{The Bayesian Low-Rank Determinantal Point Process Mixture Model}

\numberofauthors{3} 
\author{
\alignauthor
Mike Gartrell\\
       \affaddr{Microsoft}\\
       \email{mike.gartrell@acm.org}
\alignauthor
Ulrich Paquet\titlenote{Currently at Google DeepMind.} \\
       \affaddr{Microsoft} \\
       \email{ulripa@microsoft.com}
\alignauthor 
Noam Koenigstein\\
       \affaddr{Microsoft}\\
       \email{noamko@microsoft.com}
}

\makeatletter
\let\@copyrightspace\relax
\makeatother

\begin{document}

\maketitle

\begin{abstract}
Determinantal point processes (DPPs) are an elegant model for encoding
probabilities over subsets, such as shopping baskets, of a ground set, such as
an item catalog.  They are useful for a number of machine learning tasks,
including product recommendation.  DPPs are parametrized by a positive
semi-definite kernel matrix.  Recent work has shown that using a low-rank
factorization of this kernel provides remarkable scalability improvements that
open the door to training on large-scale datasets and computing online
recommendations, both of which are infeasible with standard DPP models that use
a full-rank kernel.  In this paper we present a low-rank DPP mixture model that
allows us to represent the latent structure present in observed subsets as a
mixture of a number of component low-rank DPPs, where each component DPP is
responsible for representing a portion of the observed data.
The mixture model allows us to effectively address the capacity constraints of
the low-rank DPP model. We present an efficient and scalable Markov Chain Monte
Carlo (MCMC) learning algorithm for our model that uses Gibbs sampling and
stochastic gradient Hamiltonian Monte Carlo (SGHMC).  Using an evaluation on
several real-world product recommendation datasets, we show that our low-rank
DPP mixture model provides substantially better predictive performance than is
possible with a single low-rank or full-rank DPP, and significantly better
performance than several other competing recommendation methods in many cases.
\end{abstract} 

\section{Introduction}

Online shopping activity has grown rapidly in recent years.  Central to
the online retail experience is the recommendation task of ``basket completion'',
where we seek to compute predictions for the next item that should be added to a
shopping basket, given a set of items already present in the basket.
Determinantal point processes (DPPs) offer an attractive model for basket
completion, since they jointly model \emph{set diversity} and item
\emph{quality} or \emph{popularity}.  DPPs also offer a compact parameterization
and efficient algorithms for performing inference.

A distribution over sets that models diversity is of particular interest when
recommendations are complementary.  For example, consider a shopping basket
that contains a smartphone and a SIM card.  A collaborative filtering method
based on user and item similarities, such as a matrix factorization
model~\cite{paquet2013one}, would tend to provide recommendations that are
similar to the items already present in the basket but not necessarily
complementary.  In this example, matrix factorization might recommend other
similar smartphones to complete this basket, which may not be appropriate since
the basket already contains a smartphone.  In contrast, a complementary
recommendation for this basket might be a smartphone case, rather than another
smartphone.  In this setting, DPPs would be used to learn the inherent item
diversity present within the observed sets (baskets) that users purchase, and
hence can provide such complementary recommendations.

DPPs have been used for a variety of machine learning tasks~\cite{
kulesza2010structured, kulesza2012, kwok2012priors}.  DPPs can be parameterized
by a $M \times M$ positive semi-definite $\L$ matrix, where $M$ is the size of
the item catalog.  There has been some work focused on learning DPPs from
observed data consisting of example subsets~\cite{affandi2014learning,
gartrell2016OptLowRank, gartrell2016BayesianLowRank, gillenwater2014EM,
kulesza2011learning, mariet15}, which is a challenging learning task that is
conjectured to be NP-hard~\cite{kulesza2012}.
Some of this recent work has involved learning a nonparametric full-rank $\L$
matrix~\cite{gillenwater2014EM, mariet15} that does not constrain $\L$ to take a
particular parametric form, while other recent work has involved learning a
low-rank factorization of this nonparametric $\L$
matrix~\cite{gartrell2016OptLowRank, gartrell2016BayesianLowRank}.  A low-rank
factorization of $\L$ enables substantial improvements in runtime performance
compared to a full-rank DPP model during training and when computing
predictions, on the order of 10-20x or more, with predictive performance that is
equivalent to or better than a full-rank model.

While the low-rank DPP model scales well, it has a fundamental limitation
regarding model capacity due to the nature of the low-rank factorization of
$\L$.  The low-rank DPP mixture model allows us to address these capacity
constraints. As we explain in Section~\ref{sec:background}, a $K$-rank
factorization of $\L$ has an implicit constraint on the space of possible
subsets, since it places zero probability mass on subsets (baskets) with more
than $K$ items.  When trained on a dataset containing baskets with at most $K$
items, we observe from the results in ~\cite{gartrell2016OptLowRank} that this
constraint is reasonable and that the rank-$K$ DPP provides predictive
performance that is approximately equivalent to that of the full-rank DPP
trained on the same dataset.  Therefore, in this scenario the rank-$K$ DPP can
be seen as a good approximation of the full-rank DPP.  However, we empirically
observe that the rank-$K$ DPP generally does not provide improved predictive
performance for values of $K$ greater than the size of the largest basket in the
data.  Thus, for a dataset containing baskets no larger than size $K$, there is
generally no utility in increasing the number of low-rank DPP item trait
dimensions beyond $K$, which establishes an upper bound on the capacity of the
model.  This limitation motivates us to find a way to move beyond the capacity
of the low-rank DPP model.  We present a mixture model composed of a number of
component low-rank DPPs as an effective method for addressing the capacity
constraints of the low-rank DPP.  In the low-rank DPP mixture model, each
component low-rank DPP is responsible for modeling only a \emph{subset} of the
baskets in the full dataset.  In the case of the conventional low-rank DPP
model, the low-rank DPP is responsible for modeling the \emph{entire} dataset. 
Therefore, the mixture model provides us with capacity beyond that available
with only a single low-rank DPP.

Our work makes the following contributions:
\begin{enumerate}
  \item We present a Bayesian low-rank DPP mixture model, which represents the
  latent structure present in observed subsets as a mixture of a number of
  component low-rank DPPs.  The mixture model allows us to effectively address
  the capacity constraints of the low-rank DPP model.
  \item We present an efficient and parallelizable learning algorithm for our
  model that utilizes Gibbs sampling and stochastic gradient Hamiltonian Monte
  Carlo (SGHMC).
  \item A detailed experimental evaluation on several real-world datasets shows
  that our mixture model provides significantly better predictive performance
  than existing DPP models.  Our model also provides significantly better
  predictive performance than several other state-of-the-art recommendation
  methods in many cases.
\end{enumerate}

\vspace{-1.1em}
\section{Model}
\label{sec:model}

DPPs originated in statistical mechanics~\cite{macchi1975}, where they were used
to model distributions of fermions.  Fermions are particles that obey the Pauli
exclusion principle, which indicates that no two fermions can occupy the same
quantum state.  As a result, systems of fermions exhibit a repulsion or
``anti-bunching'' effect, which is described by a DPP.   This repulsive behavior
is a key characteristic of DPPs, which makes them a capable model for diversity.
 We now proceed with some details of DPPs, including how they are defined and a
 method for efficient learning.

\subsection{Background}
\label{sec:background}

A point process is a distribution over configurations of points selected from a
ground set $\Ycal$, which are finite subsets of $\Ycal$.
In this paper we deal only with discrete DPPs, which describe a distribution
over subsets of a discrete ground set of items $\Ycal = {1, 2, \ldots, M}$,
which we also call the item catalog.  A discrete DPP on $\Ycal$ is a probability
measure $\Pcal$ on $2^{\Ycal}$ (the power set or set of all subsets of $\Ycal$),
such that for any $A \subseteq \Ycal$, the probability $\Pcal(A)$ is specified
by $\Pcal(A) \propto \det(\L_A)$.  In the context of basket completion, $\Ycal$
is the item catalog (inventory of items on sale), and $A$ is the subset of items
in a user's basket; there are $2^{|\Ycal|}$ possible baskets.
The notation $\L_A$ denotes the principal submatrix of the DPP kernel $\L$
indexed by the items in $A$, which is the restriction of $\L$ to the rows and
columns indexed by the elements of $A$: $\L_A \equiv [\L_{ij}]_{i, j \in A}$.
Intuitively, the diagonal entry $L_{ii}$ of the kernel matrix $\L$ captures the
importance or quality of item $i$, while the off-diagonal entry $L_{ij} =
L_{ji}$ measures the similarity between items $i$ and $j$.

The normalization constant for $\Pcal$ follows from the observation that
$\sum_{A' \subseteq \Ycal} \det(\L_{A'}) = \det(\L + \I)$.
The value $\det(\L_{A})$ associates a ``volume'' to basket $A$ from a geometric
viewpoint, and its probability is normalized by the volumes of all possible
baskets $A' \subseteq \Ycal$. Therefore, we have
\begin{align} \label{eq:dpp}
\Pcal(A) = \frac{\det(\L_A)}{\det(\L + \I)} \ .
\end{align}
We use a low-rank factorization of the $M \times M$ $\L$ matrix,
\begin{equation} \label{eq:Llowrank}
\L = \V \V^T \ ,
\end{equation}
for the $M \times K$ matrix $\V$, where $M$ is the number of items in the item
catalog and $K$ is the number of latent trait dimensions.  This low-rank
factorization of $\L$ leads to significant efficiency improvements compared to a
model that uses a full-rank $\L$ matrix when it comes to model learning and
computing predictions~\cite{gartrell2016OptLowRank}.  This also places an implicit
constraint on the space of subsets of $\Ycal$, since the model is restricted to
place zero probability mass on subsets with more than $K$ items (all eigenvalues
of $\L$ beyond $K$ are zero).  We see this from the observation that a sample
from a DPP will not be larger than the rank of
$\L$~\cite{gillenwater2014approx}.

\subsection{Model Specification and Learning}

\begin{figure}[t]
  \begin{center}
    \begin{tikzpicture}
  \node[obs]                               (A) {$A_n$};
  \node[latent, left=of A]                (v) {$\v_i$};
  \node[latent, left=of v]                (gamma) {$\gamma$};
  \node[const, left=of gamma]                (ab) {$a_0, b_0$};

  \edge {v} {A} ; 
  \edge {gamma} {v} ;
  \edge {ab} {gamma} ; 

  \plate {A} {(A)} {$N$} ;
  \plate {items-plate} {(v)} {$M$} ;
\end{tikzpicture}
  \end{center}
  \vspace{-1.5em}
  \caption{A graphical model for the low-rank DPP model.}
  \label{fig:DPP-graphical-model}
  \vspace{-1.0em}
\end{figure}
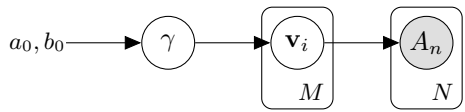

\begin{figure}[t]
  \begin{center}
    \begin{tikzpicture}
  \node[obs]                              (A) {$A_n$};
  \node[latent, left=of A]                (z) {$z_n$};
  \node[latent, left=of z]                (phi) {$\phi$};
  \node[const, left=of phi]               (alpha) {$\alpha$}; 
  \node[latent, above=2 cm of A]          (v) {$\v_{wi}$};
  \node[latent, left=of v]                (gamma) {$\gamma_w$};
  \node[const, left=of gamma]             (ab) {$a_0, b_0$};

  \edge {v} {A} ; 
  \edge {gamma} {v} ;
  \edge {ab} {gamma} ;
  \edge {z} {A} ;
  \edge {phi} {z}
  \edge {alpha} {phi} ; 

  \plate {A} {(A)(z)} {$N$} ;
  \plate {items-plate} {(v)} {$M$} ;
  \plate {mixcomponents-plate} {(v)(items-plate)(gamma)} {$W$} ;
\end{tikzpicture}
  \end{center}
  \vspace{-1.5em}
  \caption{A graphical model for the low-rank DPP mixture model.}
  \label{fig:mixture-graphical-model}
  \vspace{-1.0em}
\end{figure}
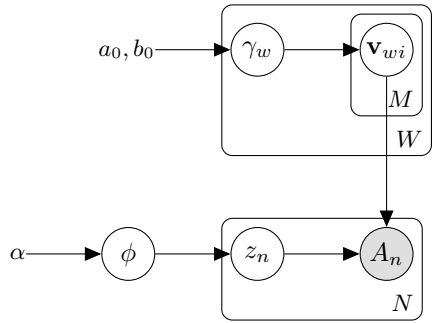

We begin by describing DPP learning and the Bayesian low-rank DPP
model~\cite{gartrell2016BayesianLowRank}, since our DPP mixture model builds on
these concepts. The DPP learning task is to fit a DPP kernel $\L$ based on a
collection of $N$ observed subsets $\Acal = \{A_1, \ldots, A_N\}$, where each
subset $A_n$ is composed of items from the item catalog $\Ycal$.  These observed
subsets in $\Acal$ constitute our training data, and our task is to infer $\L$
from $\Acal$.  The log-likelihood for seeing $\Acal$ is
\begin{align}
f(\V) & = \log \Pcal(\Acal | \V) = \sum_{n=1}^{N} \log \Pcal(A_n | \V) \\
& = \sum_{n=1}^{N} \log \det(\L_{[n]}) - N \log \det(\L + \I)
\label{eq:low-rank-DPP-likelihood}
\end{align}
where $[n]$ indexes the observed subsets in $\Acal$.
Recall from (\ref{eq:Llowrank}) that $\L = \V \V^T$.

Figure~\ref{fig:DPP-graphical-model} shows the graphical model for the Bayesian
low-rank DPP model.  We place a multivariate Gaussian prior on each item in our
model.  Our prior distribution on $\V$ is given by
\begin{align}
p(\V | \gamma) = \prod_{i = 1}^M \Ncal(\v_i ; \0, \gamma^{-1} \I)
\end{align}
where $\v_i$ is the row vector from $\V$ for item $i$, and all items $i$ share the same precision
$\gamma$.
We furthermore place a conjugate gamma prior on $\gamma$: $p(\gamma | a_0, b_0)
= \text{Gamma}(\gamma ; a_0, b_0)$.
The joint distribution over $\Acal$, $\V$ and $\gamma$, as depicted in 
Figure~\ref{fig:DPP-graphical-model}, is
\begin{align}
p(\Acal, \V, \gamma | a_0, b_0) & = \Pcal(\Acal | \V) \, p(\V | \gamma) \, p(\gamma | a_0, b_0) \ .
\end{align}

For our Bayesian low-rank DPP mixture model, we consider a mixture model
composed of $W$ component low-rank DPPs:
\begin{align}
p(\Acal_n | \Vline, \bphi) = \sum_{w = 1}^W \phi_w \Pcal(\Acal_n |
\V_w)
\end{align}
where $\Vline$ denotes the collection of $\V_w$ components for all $W$,  $\{
\V_w \}_{w=1}^W$.  This is a convex combination of the $W$ low-rank DPP components,
where the mixing weights $\phi_w$ satisfy $0 \leq \phi_w \leq 1$ and $\sum_{w =
1}^W \phi_w = 1$.

If we use the formulation of the Bayesian low-rank DPP model described above,
then we have the following model specification for a low-rank DPP mixture model:
\begin{align}
W & = \text{number of mixture components} \nonumber \\
N & = \text{number of observed sets} \nonumber \\
\Ycal & = \text{item catalog} \nonumber \\
M & = \text{number of items in } \Ycal \nonumber \\
K & = \text{\parbox{6cm}{number of item trait dimensions for the component
low-rank DPPs}} \nonumber\\
A_{n = 1 \ldots N} & = \text{observation } n \nonumber \\
\phi_{w = 1 \ldots W}, \bphi & = \text{\parbox{6cm}{$W$-dimensional vector
composed of all the individual $\phi_w$; must sum to 1}} \nonumber \\
\V_{w = 1 \ldots W} & = \text{\parbox{6cm}{item trait matrix for low-rank DPP
component $w$}} \nonumber \\
\z_{n = 1 \ldots N} & = \text{component of observation } n \nonumber \\
a_0, b_0, \alpha & = \text{shared hyperparameters} \nonumber \\
\bphi & \sim \text{Symmetric-Dirichlet}_W(\alpha) \nonumber \\
\z_{n = 1 \ldots N} & \sim \text{Categorical}(\bphi) \nonumber \\
\gamma_w & \sim \text{Gamma}(\gamma_w; a, b) \nonumber \\
\V_w & \sim \prod_{i = 1}^M \Ncal(\v_{wi} | 0, \gamma_w^{-1} \I) \nonumber \\ 
A_{n = 1 \ldots N} & \sim \Pcal(\V_{z_n})
\end{align}
where $\v_{wi}$ is the row vector from $\V_w$ for mixture component $w$ and item
$i$.  We set $\alpha = \frac{1}{W}$.  With this specification for the Dirichlet
prior, for $W$ set to a conservative upper bound on the number of occupied
mixture components, the model approximates a Dirichlet process mixture
model~\cite{ishwaran2002}, where redundant mixture components are automatically
given zero weight during learning~\cite{rousseau2011}.  We find $W = 100$ works
well for the datasets we evaluated.  Figure~\ref{fig:mixture-graphical-model}
shows the graphical model for our low-rank DPP mixture model.  To draw samples
from the posterior $p(\V_w, \bphi | \Acal,
\alpha, \gamma_w)$, for $w = 1 \ldots W$, we use the Gibbs sampling
algorithm shown in Algorithm~\ref{alg:sampling}.

We can write $\z_n$ as a $W$-dimensional binary random variable having a
1-of-$W$ representation in which a particular element $z_{nw}$ is equal to 1 if
the $n$th observation is drawn from the $w$th mixture component, and all other
elements are equal to 0.  $\Z$ is a $N \times W$ matrix, where $\z_n$ is a row
vector from this matrix.  

We can write the joint likelihood function for this model as
\begin{align}
p(\Acal, \Z | \Vline, \bphi) & = \prod_{n = 1}^N \prod_{w = 1}^W \phi_w^{z_{nw}}
\Pcal(A_n | \V_w)^{z_{nw}} \ .
\end{align}

The conditional distribution $p(\Z | \Acal, \Vline, \bphi)$ is
\begin{align}
p(\Z | \Acal, \Vline, \bphi) & \propto \prod_{n = 1}^N \prod_{w = 1}^W \left(
\phi_w \Pcal(A_n | \V_w) \right)^{z_{nw}} \ .
\end{align}

Therefore, we can draw samples for each discrete assignment $\z_n$ from
\begin{align}
p(\z_{n} | A_n, \Vline, \bphi)  \propto \exp \Bigg\{ \sum_{w = 1}^W z_{nw}
\left( \log \phi_w + f(A_n | \V_w) \right) \Bigg\} \nonumber
\end{align}
directly, as each $p(\z_{n} | A_n, \V_w, \bphi)$ is an independent categorical
distribution.  Note that $f(A_n | \V_w)$ is the DPP log-likelihood
(Equation~\ref{eq:low-rank-DPP-likelihood}) for a particular observation $A_n$.
Subtract the maximum $m = \max_w (\log \phi_w + f(A_n | \V_w))$ everywhere,
\begin{align}
p(z_{nw} = 1 | \Acal, \Vline, \bphi) = \frac{ \exp \big\{ \log \phi_w + f(A_n
| \V_w) - m \big\} }{ \sum_{w' = 1}^W \exp \big\{ \log \phi_{w'} + f(A_n
| \V_{w'}) - m \big\} } \ ,
\label{eq:z_{nw}}
\end{align}
so that the biggest exponentiated component is $\exp(0)$, and normalize to sum
to one. Then draw the discrete index for which $z_{nw} = 1$.

The conditional distribution $p(\bphi | \Acal, \Z, \alpha)$ is
\begin{align}
p(\bphi | \Acal, \Z, \alpha) & \propto \prod_{n = 1}^N \text{Cat}(\z_n | \bphi)
\text{ Symmetric-Dirichlet}(\bphi | \alpha) \nonumber \\
& = \text{Dirichlet} \left( \bphi | \alpha + \sum_{n = 1}^N
z_{n1}, \ldots, \alpha + \sum_{n = 1}^N z_{nw} \right)
\label{eq:phi}
\end{align}
and is sampled in Line~\ref{line:phi} in Algorithm~\ref{alg:sampling}.

The conditional distribution $p(\gamma_w | \V_w, a_0, b_0)$ is
\begin{align}
\gamma_w | \V_w, a_0, b_0 & \sim \text{Gamma}(\gamma_w ; a, b) \nonumber \\
 a & = a_{0} + \frac{M K}{2} \nonumber \\ 
 b & = b_{0} + \frac{1}{2} \sum_{i=1}^M \| \v_{wi} \|^2 \ ,
 \label{eq:gamma} 
\end{align}
and is sampled in Line~\ref{line:gamma} in Algorithm~\ref{alg:sampling}.

The conditional distribution $p(\V_w | \Acal, \Z, \gamma_w)$ is
\begin{align}
p(\V_w | \Acal, \Z, \gamma_w) & \propto \prod_{n = 1}^N \Bigg( \exp \big(
f(A_n | \V_w) \big) \Bigg. \nonumber \\
&\qquad \Bigg. {} \prod_{i = 1}^M \Ncal(\v_{wi} | \0, \gamma_w^{-1} \I)
\Bigg)^{z_{nw}} \nonumber  \\
\log p(\V_w | \Acal, \Z, \gamma_w) & = \sum_{n = 1}^N z_{nw} \Bigg( 
f(A_n | \V_w) \big. \nonumber \\
&\qquad \left. {} + \sum_{i = 1}^M \frac{1}{2} \v_{wi}^T (\gamma_w \I) \v_{wi}
\right) + \mathrm{const}
\label{eq:V-conditional}
\end{align}
where $\const$ indicates an additive
constant independent of $\V_w$.  In the following section we show how to sample
from $p(\V_w | \Acal, \Z, \gamma_w)$.

\subsection{Learning Algorithm}

We estimate the conditional distribution $p(\V_w | \Acal, \Z, \gamma_w)$
using stochastic gradient Hamiltonian Monte Carlo (SG\hyp{}HMC)~\cite{chen2014}.
Hamiltonian Monte Carlo (HMC)~\cite{duane1987, neal2011} is a Markov chain Monte
Carlo (MCMC) method that uses the gradient of the log-density of the target
distribution to efficiently explore the state space of the target.  HMC defines
a Hamiltonian function, an idea borrowed from physics, in terms of the target
distribution that we wish to sample from.  The Hamiltonian function has a
potential energy term, corresponding to the target distribution, and a kinetic
energy term, defined in terms of auxiliary momentum variables.  By updating the
momentum variables using the gradient of the log-density of the target
distribution, we simulate a Hamiltonian dynamical system that enables proposals
of distant states, thus allowing HMC to move rapidly through the state space of
the target.  We cannot simulate directly from the continuous Hamiltonian
dynamics, so HMC uses a discretization of this continuous system composed of a
number of ``leapfrog steps''.

HMC requires computation of the gradient of the log-density of the target
distribution over all training instances with each iteration of the algorithm,
which is expensive or infeasible for large datasets or a complex target.
SGHMC addresses this issue by using stochastic gradients that are computed on
minibatches, where each minibatch is composed of training instances sampled
uniformly at random from the full training set.  SGHMC adds a friction term to
the momentum update, which counteracts the effects of noise from the stochastic
gradients.  The estimates computed by SGHMC samples are not unbiased
(notice that there is no accept-reject step in Algorithm \ref{alg:sampling}, and the
distribution that is sampled from is different from the true posterior), but due
to the effectively faster mixing, we are able to efficiently train our Bayesian
low-rank DPP mixture model on large-scale datasets.

Since we learn $p(\V_w | \Acal, \Z, \gamma_w)$ by SGHMC, we need to efficiently compute
the gradient of the log-density for this distribution.  We begin by computing
the gradient of the low-rank DPP log-likelihood, $\partial f / \partial \V$,
which will be a $M \times K$ matrix. For $i \in 1, \ldots, M$ and $k \in 1,
\ldots, K$, we need a matrix of scalar derivatives, 
$ \left\{ \frac{\partial f}{\partial \V} \right\}_{ik} =
\frac{\partial f}{\partial v_{ik}} \ .
$
Taking the derivative of each term of the log-likelihood, we have 
\begin{align}
\frac{\partial f}{\partial v_{ik}}
& =
\sum_{n : i \in [n]} \frac{\partial}{\partial v_{ik}} \Big( \log \det(\L_{[n]})
\Big) - N \frac{\partial}{\partial v_{ik}} \Big( \log \det(\L + \I) \Big)
\nonumber
\\
& =
\sum_{n : i \in [n]} \trace \left( \L_{[n]}^{-1}  \frac{\partial
\L_{[n]}}{\partial v_{ik}} \right) - N \trace \left( (\L + \I)^{-1} 
\frac{\partial (\L + \I)}{\partial v_{ik}} \right) \ .
\end{align}

Examining the first term of the derivative, we see that 
\begin{align}
\trace \left( \L_{[n]}^{-1} \frac{\partial \L_{[n]}}{\partial v_{ik}} \right) =
\a_{i} \cdot \v_k + \sum_{j = 1}^{M} a_{ji} v_{jk} \ ,
\end{align}
where $\a_i$ denotes row $i$ of the matrix $\A = \L_{[n]}^{-1}$ and $\v_k$
denotes column $k$ of $\V_{[n]}$.
Note that $\L_{[n]} = \V_{[n]} \V_{[n]}^T$.
Computing $\A$ is a relatively inexpensive operation, since the number
of items in each training instance $A_n$ is generally small for many
recommendation applications.

For the second term of the derivative, we see that  
\begin{align}
\trace \left( (\L + \I)^{-1}  \frac{\partial (\L + \I)}{\partial v_{ik}} \right)
= \b_{i} \cdot \v_k + \sum_{j = 1}^{M} b_{ji} v_{jk}
\end{align}
where $\b_i$ denotes row $i$ of the matrix $\B = \I_m - \V(\I_k + \V^T \V)^{-1}
\V^T$.  Computing $\B$ is a relatively inexpensive operation, since we are
inverting a $K \times K$ matrix with cost $O(K^3)$, and $K$ (the number of
latent trait dimensions) is usually set to a small value.

We now proceed with computing the gradient of the log-density for $p(\V_w |
\Acal, \Z, \gamma_w)$ shown in Equation~\ref{eq:V-conditional}.  Looking
at one component of this gradient, we have:
\begin{align}
\frac{\partial \log p(\V_w | \Acal, \Z, \gamma_w)}{\partial v_{wik}} &  =
\sum_{n = 1}^N z_{nw} \left( \frac{\partial f_w(\Acal_n)}{\partial
v_{wik}} + \gamma_w v_{wik} \right)
\label{eq:gradient}
\end{align}

\begin{algorithm}[t]
\caption{Sampling algorithm for learning the posterior $p(\V_w, \bphi | \Acal,
\alpha, \gamma_w)$ for $w = 1 \ldots W$}
\label{alg:sampling}
\begin{algorithmic}[1]
\STATE \textbf{initialize} $\Vline$ randomly, $\Z$ and $\bphi$ by drawing from
priors, $\R = 0$ \STATE $\mathsf{samples} := \{ \}$
\REPEAT
\STATE randomly select minibatch $\tilde{\Acal}$ from $\Acal$
\STATE sample $p(\Z | \tilde{\Acal}, \Vline, \bphi)$ according to
(\ref{eq:z_{nw}}) \label{line:Z} 
\STATE sample $p(\bphi | \tilde{\Acal}, \Z, \alpha)$ according to (\ref{eq:phi})
\label{line:phi}
\FOR{mixtureComponents $w = 1, \ldots, W$} \label{line:for-components}
	\STATE sample $p(\gamma_w | \V_w, a_0, b_0)$ according to (\ref{eq:gamma})
	\label{line:gamma} 
	\STATE \emph{/\!/ approximately sample $\V_w | \Acal, \Z, \gamma_w$:}
	\FOR{leapfrogSteps $j = 1, \ldots, L$}
		\STATE $\R := \eta \nabla \tilde{U}(\V_w) - \beta \R + \Ncal(0, 2 \alpha  
		\eta)$ \STATE $\V_w := \V_w + \R$
	\ENDFOR
\ENDFOR
\STATE $\mathsf{samples} := \{ \mathsf{samples}, \Vline, \bphi \}$
\UNTIL{sufficient samples have been taken}

\end{algorithmic}
\end{algorithm}

Our sampling algorithm that utilizes SGHMC is shown in
Algorithm~\ref{alg:sampling}.  In this algorithm, all samples that are collected
from conditional distributions that involve $\Acal$ are computed on a minibatch
$\tilde{\Acal}$, so that we preserve the efficiency gains provided by use of
minibatches throughout the sampling algorithm.  $\nabla \tilde{U}(\V)$ is a
noisy estimate of the log-density gradient of $p(\V_w | \Acal, \Z, \bphi,
\gamma_w)$ computed for $\tilde{\Acal}$ using Equation~\ref{eq:gradient}, $\eta
> 0$ is the learning rate, $\beta \in [0, 1]$ is the momentum coefficient, and
$\R$ is the auxiliary momentum variable.  We find that setting $\beta = 0.01$
and $\eta = 1.0 \times 10^{-5}$ or $\eta = 1.0 \times 10^{-6}$, with a minibatch
size of 5000 instances, works well for the datasets we tested.  The \textbf{for}
loop on Line~\ref{line:for-components} of this algorithm may be easily
parallelized, since there are no dependencies between the parameters $\V_w$ and
$\gamma_w$ for each mixture component.

\subsection{Predictions}

We compute singleton next-item predictions, given a set of observed
items.  An example of this class of problem is ``basket completion'', where we
seek to compute predictions for the next item that should be added to shopping
basket, given a set of items already present in the basket. 

We use a mixture of $k$-DPPs to compute next-item predictions.  A $k$-DPP is a
distribution over all subsets $A \in \Ycal$ with cardinality $k$, where $\Ycal$ is the ground
set, or the set of all items in the item catalog.  Next item predictions are
done via a conditional density.  We compute the probability given the observed
basket $A$, consisting of $k$ items.  For each possible item to be recommended,
given the basket, the basket is enlarged with the new item to $k+1$ items.
For the new item, we determine the probability of the new set of $k+1$ items,
given that $k$ items are already in the basket, using a Monte Carlo estimate from the samples.
Ignoring burn-in samples, and letting $s$ index the $S$ remaining $\Vline^{(s)}$
and $\bphi^{(s)}$ samples in $\mathsf{samples}$,
\begin{align}
p(A_{+1} | A, \Acal, a_0, b_0, \alpha) & = \iint \Pcal(A_{+1} | A, \Vline,
\bphi)\nonumber \\
&\qquad {}p(\Vline, \bphi | \Acal, \alpha, a_0, b_0)  \, d \Vline \, d\bphi
\nonumber \\
& \approx \frac{1}{S} \sum_{s=1}^S \sum_{w=1}^W \phi_w^{(s)} \Pcal(A_{+1} | A,
\V_w^{(s)})
\end{align}
where $A_{+1}$ indicates set $A$ enlarged to contain a new item $b$ from the
catalog $\Ycal$.  The samples in $\Vline$ and $\bphi$ implicitly marginalize out
$\gamma_w$ and $\Z$ from the posterior density.  We run the sampler to generate
2,000 samples, and discard the first 1,800 samples as burn-in.
From~\cite{gartrell2016OptLowRank}, we see that for the DPP for a single
component from the mixture model, the conditional probability for an item
$b$ in the singleton set $B$, given the appearance of items in $A$, is
\begin{align}
\Pcal(A_{+1} = A \cup B | A) 
& = \frac{L^A_{bb}}{e_1(\lambda^A_1, \lambda^A_2, \ldots, \lambda^A_N)}
\end{align}
where $L^A_{bb}$ denotes diagonal element $bb$ from the $k$-DPP kernel matrix
conditioned on $A$, $\L^A$; $\lambda^A_1, \allowbreak \lambda^A_2, \allowbreak
\ldots, \allowbreak \lambda^A_N$ are the eigenvalues of $\L^A$; and
$e_1(\lambda^A_1, \allowbreak \lambda^A_2, \allowbreak \ldots, \allowbreak
\lambda^A_N)$ is the first elementary symmetric polynomial on these eigenvalues.
 See~\cite{gartrell2016OptLowRank} for full details on how to efficiently compute
conditional densities for a $k$-DPP given an observed basket.

\section{Evaluation}
\label{sec:evaluation}

In this section we evaluate the performance of the Bayesian low-rank DPP mixture
model on on the task of basket completion for several real-world datasets.  We
compare to several competing recommendation methods, including an
optimization\hyp{}based low-rank DPP~\cite{gartrell2016OptLowRank}, a Bayesian
low-rank DPP~\cite{gartrell2016BayesianLowRank}, a full-rank
DPP~\cite{mariet15}, and two matrix factorization models~\cite{gopalan2015,
paquet2013one}, and find that our approach provides better predictive
performance in many cases.

We formulate the basket-completion task as follows.  Let $A_{\mathrm{test}}$ be a subset of
$n\geq2$ co-purchased items (i.e, a basket) from the test-set.  In order to
evaluate the basket completion task, we pick an item $i \in A_{\mathrm{test}}$ at random and
remove it from $A_{\mathrm{test}}$. We denote the remaining set as $A_{\mathrm{test}-1}$. Formally,
$A_{\mathrm{test}-1} = A_{\mathrm{test}} \diagdown \, \{i\}$. Given a ground set of possible items
$\Ycal=1,2,...,M$, we define the candidates set $\Ccal$ as the set of all items
except those already in $A_{\mathrm{test} - 1}$; i.e., $\Ccal = \Ycal \diagdown \,A_{\mathrm{test}-1}$.
Our goal is to identify the missing item $i$ from all other items in $\Ccal$. 

\subsection{Datasets}
Our experiments are based on several datasets:
\begin{enumerate}
	\item \textbf{Amazon Baby Registries -} 
Amazon\footnote{\url{www.amazon.com}} is one of the world's leading online
retail stores.  The Amazon Baby Registries dataset~\cite{gillenwater2014EM} is a
public dataset consisting of 111,006 registries or ``baskets'' of baby products.
The choice of this dataset was motivated by the fact that it has been used by
several prominent DPP studies~\cite{gartrell2016OptLowRank, gillenwater2014EM,mariet15}. 
The registries are collected from 15 different categories (such as ``feeding'',
``diapers'', ``toys'', etc.), and the items in each category are disjoint.
We maintain consistency with prior work by evaluating each of its categories
separately using a random split of 70\% of the data for training and 30\% for
testing.
	
In addition to the above evaluation, we also constructed a dataset composing of
the concatenation of the three most popular categories: apparel, diaper, and
feeding.  This three-category dataset allows us to simulate data that could be
observed for department stores that offer a wide range of items in different
product categories.  Its construction is deliberate, and concatenates three
disjoint subgraphs of basket-item purchase patterns. This dataset serves to
highlight differences between DPPs and models based on matrix factorization
(MF).  Collaborative filtering-based MF models -- which model each basket and
item with a latent vector -- will perform poorly for this dataset, as the latent
trait vectors of baskets and items in one subgraph could be arbitrarily rotated,
without affecting the likelihood or predictive error in any of the other
subgraphs.  MF models are invariant to global rotations of the embedded trait
vectors. However, for the concatenated dataset, these models are also invariant
to arbitrary rotations of vectors in each disjoint subgraph, as there are no
shared observations between the three categories.  A global ranking based on
inner products could then be arbitrarily affected by the basket and item
embeddings arising from each subgraph.

	\item \textbf{MS Store -}			
This dataset is based on data from Microsoft's web-based
store~\footnote{\url{microsoftstore.com}}.  The dataset is composed of 243,147
baskets consisting of commonly purchased items from a catalog of 2097 different
hardware and software products.  We randomly sampled of 80\% of the data for
training and kept the remaining 20\% for testing.
\end{enumerate}

Since we are interested in the basket completion task, which requires baskets
containing at least two items, we remove all baskets containing only one item
from each dataset before splitting the data into training and test sets.

\subsection{Competing Methods}
We evaluate against several baselines:

\begin{enumerate}
  	\item \textbf{Full-rank DPP -}
This DPP model is parameterized by a full-rank $\L$ matrix, and uses a
fixed-point optimization algorithm called Picard iteration~\cite{mariet15}
for learning $\L$.  As described in~\cite{gartrell2016OptLowRank}, a full-rank 
DPP does not scale well to datasets containing large item catalogs during
training or when computing predictions.
	\item \textbf{Low-rank DPP trained using stochastic gradient ascent (SGA) -}
This DPP model~\cite{gartrell2016OptLowRank} is parameterized by a low-rank $\L$ matrix
that is factorized using a $\V$ matrix composed of latent item trait vectors,
and has a likelihood function identical to the Bayesian low-rank DPP.  This
optimization-based model is trained using stochastic gradient ascent, and uses
regularization based on item popularity.  We selected the regularization
hyperparameter for this model using a line search performed with a validation
set.  Recall from Section~\ref{sec:background} that a low-rank DPP places zero
probability mass on subsets with more than $K$ items, where $K$ is the number of
trait dimensions in $\V$ or the rank of $\L$.  With this constraint in mind, we
set $K$ to the size of the largest observed basket in the data when training
this model.
	\item \textbf{Bayesian low-rank DPP -}
This DPP model~\cite{gartrell2016BayesianLowRank} has a low-rank DPP likelihood
function, as shown in Eq.~\ref{eq:low-rank-DPP-likelihood}.  A multivariate
Gaussian prior is placed on each item trait vector in this model.  Since the
posterior in this model is a not a low-rank DPP, but instead a non-convex
combination of the prior and likelihood, we may see some improvement in
predictive performance for values of $K$, the number of trait dimensions, larger
than the size of the largest observed basket in the data.  We use the
hyperparameter settings
described in~\cite{gartrell2016BayesianLowRank}.
	\item \textbf{Poisson Factorization (PF) - } 
Poisson factorization (PF) is a prominent variant of matrix factorization
designed specifically for implicit ratings~\cite{gopalan2015}.
The likelihood of the PF model is based on the Poisson distribution, which is
useful with implicit datasets (e.g. datasets based on click or purchase events).
The evaluations in this paper are based on the publicly available
implementation\footnote{Note that~\cite{PFimpl} is actually an implementation of
PF with a social component, which was disabled in the course of our evaluations
since the data does not include a social graph.} from~\cite{PFimpl}.
In PF, Gamma priors are placed on the trait vectors. Following~\cite{chaney2015,
gopalan2015}, we set the Gamma shape and rate hyperparameters to 0.3.
	\item \textbf{Reco Matrix Factorization (RecoMF) - }
RecoMF is a matrix factorization model powering the Xbox Live recommendation
system~\cite{paquet2013one}.  The likelihood term of RecoMF uses a sigmoid
function to model the odds of a user liking or disliking an item in the dataset.
Unlike PF, RecoMF requires the generation of synthetic negative training
instances, and uses a scheme for sampling negatives based on popularity.  RecoMF
places Gaussian priors on the trait vectors, and gamma hyperpriors on each.  We
use the hyperparameter settings described in~\cite{paquet2013one}, which have
been found to provide good performance for implicit recommendation data.
\end{enumerate}

We use a flexible prior in our Bayesian low-rank
DPP mixture model by setting $a_{0} = \sqrt{K}$ and $b_{0} = 1$, and find that
the model is not sensitive to these settings.

The matrix-factorization models are parameterized in terms of users and items. 
Since we have no explicit users in our data, we construct ``virtual'' users from
the contents of each basket for the purposes of our evaluation, where a new user
$u_m$ is constructed for each basket $b_m$.  Therefore, the set of items that
$u_m$ has purchased is simply the contents of $b_m$. 

\begin{figure*}[t!]
	\centering
	\includegraphics[width=0.72\textwidth]{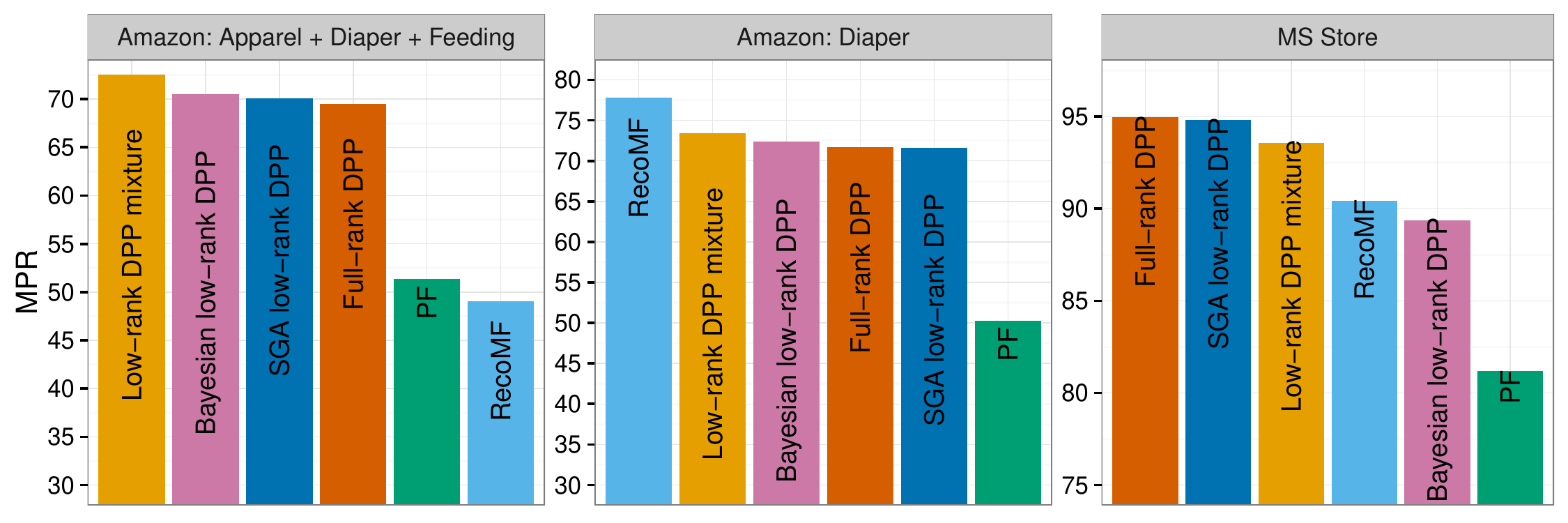}
	\vspace{-1.2em}
	\caption{Mean Percentile Rank (MPR)}
	\label{fig:MPR}
	\vspace{-0.6em}
\end{figure*}

\begin{figure*}[t!] 
	\centering
	\includegraphics[width=0.72\textwidth]{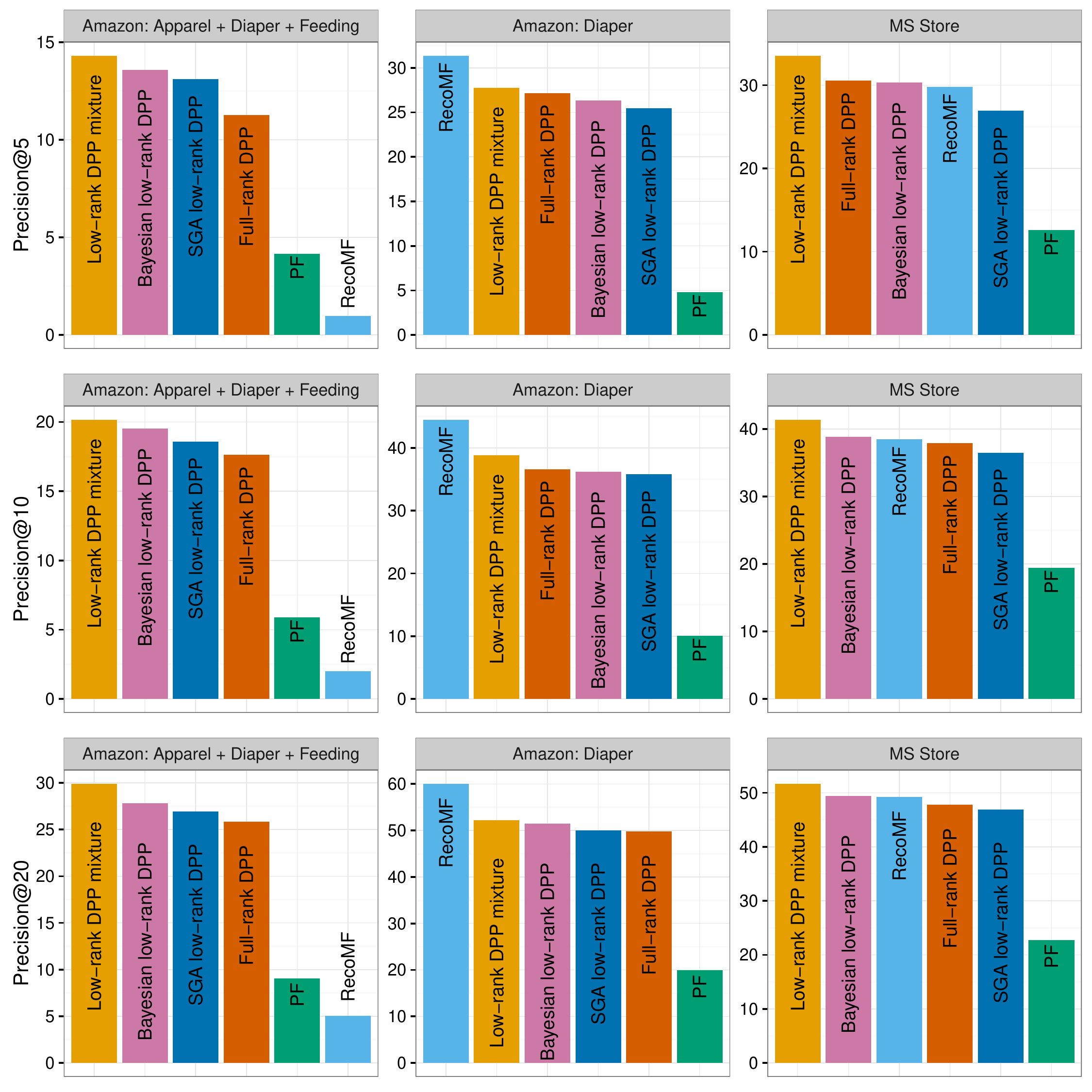}
	\vspace{-1.2em}
	\caption{Precision@$k$}
	\label{fig:precisionAtK}
	\vspace{-0.9em}
\end{figure*}

\subsection{Metrics}
In the following evaluation we consider three measures:
\begin{enumerate}
	\item \textbf{Mean Percentile Rank (MPR) - }
To compute this metric we rank the items according to their probabilities of
completing the missing set $Y_{n-1}$.  Namely, given an item $i$ from the
candidates set $\Ccal$, we denote by $p_i$ the probability $P\Large(Y_n \cup
\{i\} |Y_{n-1}\Large)$.  The Percentile Rank (PR) of the missing item $j$ is
defined by
$
\text{PR}_j = \frac{\sum_{j'\in \Ccal}\mathbb{I}\Large( p_j \ge
p_{j'}\Large)}{|\Ccal|} \times 100\% \nonumber
$,
where $\mathbb{I}\Large( \cdot \Large)$ is an indicator function and $|\Ccal|$
is the number of items in the candidates set.  The Mean Percentile Rank (MPR) is
the average PR of all the instances in the test-set:
$
\text{MPR} = \frac{\sum_{t \in \Tcal} \text{PR}_t}{|\Tcal|} \nonumber
$,
where $\Tcal$ is the set of test instances.  MPR is a recall-oriented metric
commonly used in studies that involve implicit recommendation data~\cite{hu2008,
li2010}.  $\text{MPR} = 100$ always places the held-out item for the test
instance at the head of the ranked list of predictions, while $\text{MPR} = 50$
is equivalent to random selection.
	\item \textbf{Precision@$k$ - }
We define this metric as 
$
\text{precision@}k = \frac{\sum_{t \in \Tcal} \mathbb{I}[\text{rank}_t \leq
k]}{|\Tcal|}
\nonumber
$,
where $\text{rank}_t$ is the predicted rank of the held-out item for test
instance $t$.  In other words, precision@$k$ is the fraction of instances in the
test set for which the predicted rank of the held-out item falls within the top
$k$ predictions.
	\item \textbf{Popularity-weighted precision@$k$ - }
Datasets used to evaluate recommendation systems typically
contain a popularity bias~\cite{steck2011}, where users are more likely to
provide feedback on popular items.  Due to this popularity bias, conventional
metrics such as MPR and precision@$k$ are typically biased toward popular items.
Using ideas from~\cite{steck2011}, we this metric as
$
\text{popularity-weighted precision@}k = \nonumber \\ 
\frac{\sum_{t \in \Tcal} w_t \mathbb{I}[\text{rank}_t \leq k]}{\sum_{t \in
\Tcal} w_t} \nonumber
$,
where $w_t$ is the weight assigned to the held-out item for test instance $t$,
defined as $w_t \propto \frac{1}{C(t)^\beta} \nonumber$, where $C(t)$ is the
number of occurrences of the held-out item for test instance $t$ in the training
data, and $\beta \in [0, 1]$.  The weights are normalized, so that $\sum_{j \in
\Ycal} w_j = 1$.  This popularity-weighted precision@$k$ measure assumes that
item popularity follows a power-law.  By assigning more weight to less popular
items, for $\beta > 0$, this measure serves to bias precision@$k$ towards less
popular items.  For $\beta = 0$, we obtain the conventional precision@$k$
measure.
\end{enumerate}

\subsection{Predictive Performance}

Figures~\ref{fig:MPR} and~\ref{fig:precisionAtK} show the performance of each
method and dataset for the MPR and precision@$k$ metrics.  For these plots we
selected the value of $K$, the number of trait dimensions, that provided the
best performance on the precision@5 metric for each dataset.  For the Amazon
datasets, we use $K = 30$ for the models that utilize low-rank DPPs and $K = 40$
for the RecoMF and PF models.  For the MS Store dataset, we use $K = 60$ for the
Bayesian low-rank DPP and low-rank DPP mixture models, $K = 15$ for the SGA
low-rank DPP model (the size of the largest observed basket in this dataset is
15 items), $K = 40$ for RecoMF, and $K = 90$ for PF.

We see that the low-rank DPP mixture model outperforms all other DPP models on
precision@$k$ for all datasets, often by a sizeable margin.  With the
exception of the Amazon Diaper dataset, the low-rank DPP mixture
model also provides significantly better precision@$k$ performance than
competing non-DPP models.  The low-rank DPP mixture model also provides
consistently high MPR performance on all datasets.  We attribute the strong
predictive performance of the low-rank DPP mixture model to effective use of the
increased capacity of this model afforded by the mixture components.

\begin{figure}[t!]
	\includegraphics[width=0.99\columnwidth]{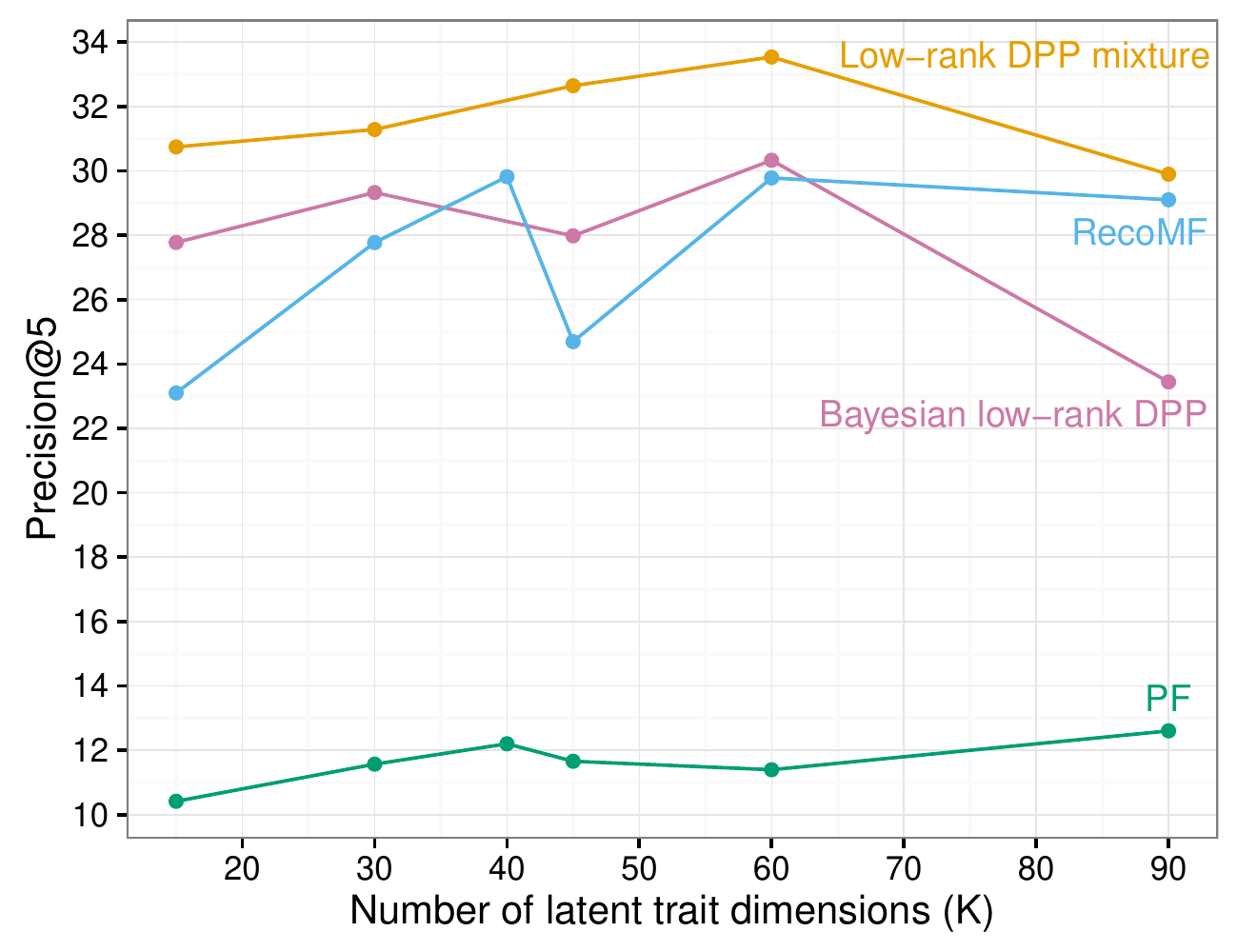}
	\vspace{-1.2em}
	\caption{Precision@5 performance of the low-rank DPP mixture model, Bayesian
	low-rank DPP, RecoMF, and PF on the MS Store dataset as a function of the
	number of latent trait dimensions $K$.}
	\label{fig:precAt5VsK}
	\vspace{-1.0em}
\end{figure}

In Figure~\ref{fig:precAt5VsK} we examine the performance of several models on
precision@5 for the MS Store dataset as a function of the number of trait
dimensions, $K$.  We see that at even for $K = 15$ the low-rank DPP mixture
model is competitive with all other models, and at $K = 60$ the mixture model
significantly outperforms all other models, with a relative improvement of
10.6\% over the score of the second-best model on this metric. 

\begin{figure}[t!]
	\includegraphics[width=0.99\columnwidth]{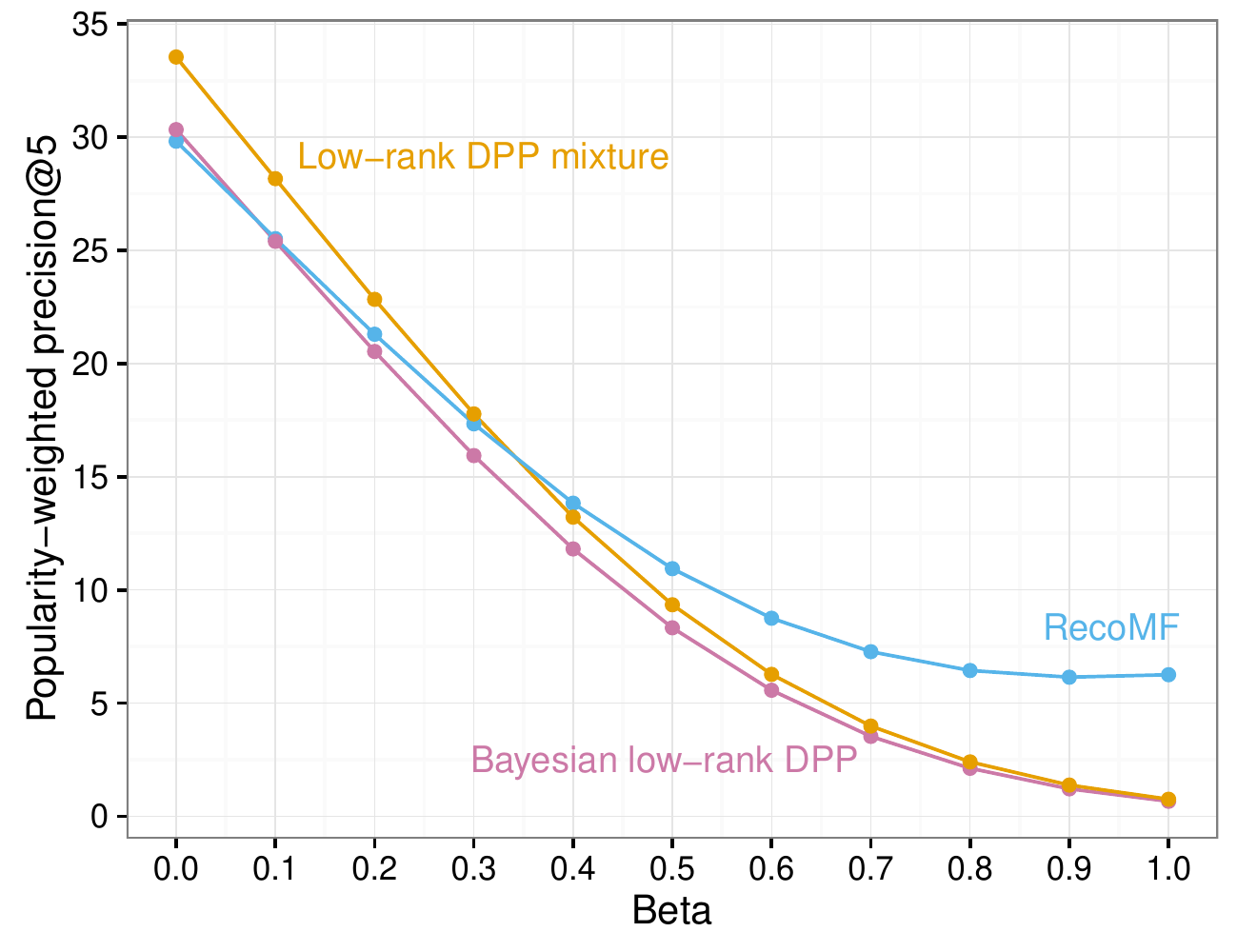}
	\vspace{-1.2em}
	\caption{Popularity-weighted precision@5 performance of the low-rank DPP
	mixture model, Bayesian low-rank DPP, and RecoMF the MS Store
	dataset as a function of the $\beta$ parameter.}
	\label{fig:popWeightedPrecAt5VsBeta}
	\vspace{-1.2em}
\end{figure}

Figure~\ref{fig:popWeightedPrecAt5VsBeta} shows how the popularity-weighted
precision@5 performance of three of the best performing models on the MS Store
dataset varies as we adjust the value of the $\beta$ parameter used in the
popularity-weighted precision@$k$ metric.  Recall that $\beta$ controls the
weight assigned to less popular items, and that for $\beta = 0$ we obtain the
standard precision@$k$ metric.  For values of $\beta \leq 0.4$ the low-rank DPP
mixture model provides leading or competitive performance, which shows that the
increased model capacity provided by the mixture model allows for improved
recommendations for less popular items in addition to popular items. For larger
values of $\beta$, RecoMF provides the best performance.  This behavior may be
explained by the scheme for sampling negatives by popularity in
RecoMF~\cite{paquet2013one}, which tends to improve recommendations for less
popular items.  The optimal setting of $\beta$ for approximating actual user
satisfaction in a real-world application will depend on the particular
application and user population.  A small-scale user study conducted with the
Netflix data~\cite{steck2011} found that $\beta = 0$ and $\beta = 0.33$ best
approximated user preferences for the application in that study.

\section{Related Work}
\label{sec:related-work}

Several algorithms for learning the DPP kernel matrix from observed data have
been proposed~\cite{affandi2014learning, gillenwater2014EM, mariet15}.
Ref.~\cite{gillenwater2014EM} presented one of the first methods for learning a
non\hyp{}parametric form of the full-rank kernel matrix, which involves an
expectation\hyp{}maximization (EM) algorithm.  In~\cite{mariet15}, a fixed-point
optimization algorithm for full-rank DPP learning is described, called Picard
iteration. Picard iteration has the advantage of being simple to implement and
performing much faster than EM during training.
Ref.~\cite{gartrell2016OptLowRank} shows that a low-rank DPP model can be
trained far more quickly than Picard iteration and therefore EM, while enabling
much faster computation of predictions than is possible with any full-rank DPP
model.  A Bayesian low-rank DPP model is presented
in~\cite{gartrell2016BayesianLowRank}, which provides robust regularization and
does not require expensive hyperparameter tuning.  The Bayesian low-rank DPP is
the fundamental building block for our low-rank DPP mixture model. 

Mixture models have been well known in the machine learning and statistics
communities for some time.  Ref.~\cite{dempster1977} describes an application of
expectation maximization (EM) to mixture models.  A variational Bayes
implementation of mixture models is described in~\cite{bishop2006},
while~\cite{minka2001} describes the use of expectation propagation for mixture
models.  A Bayesian infinite Gaussian mixture model is described
in~\cite{rasmussen1999}, where an efficient MCMC algorithm is used for
inference.  Ref.~\cite{si2003} presents a mixture model for collaborative
filtering, where user and items are clustered, while allowing each user and item
to belong to multiple clusters to account for diversity in user interests and
item aspects.

A number of approaches to the basket completion problem that we focus on in this
paper have been proposed. Ref.~\cite{mild2003} describes a
user-neighborhood-based collaborative filtering method, which uses rating data
in the form of binary purchases to compute the similarity between users, and
then generates a purchase prediction for a user and item by computing a weighted
average of the binary ratings for that item.  A technique that uses logistic
regression to predict if a user will purchase an item based on binary purchase
scores obtained from market basket data is described in~\cite{lee2005}.
Additionally, other collaborative filtering approaches could be applied to the
basket completion problem, such as the one-class matrix factorization model
in~\cite{paquet2013one} and Poisson factorization~\cite{gopalan2015}, as we
illustrate in this paper.

\section{Conclusions}
\label{sec:conclusions}

We have presented a Bayesian low-rank DPP mixture model that represents observed
subsets as a mixture of several component low-rank DPPs.  Through the use of a
stochastic MCMC learning algorithm that operates on minibatches of training
data, we are able to avoid costly updates that would require full passes through
the entire dataset as required by conventional MCMC learning algorithms. 
With an extensive evaluation on several real-world shopping basket datasets, we
have shown that our mixture model effectively addresses the capacity constraints
of conventional DPP models and provides significantly better predictive
performance than these models.  Our model also significantly outperforms
state-of-the-art methods based on matrix factorization in many cases.

\newpage

\section{Acknowledgements}
\label{sec:acknowledgements}

We thank Gal Lavee and Shay Ben Elazar for many helpful discussions.  We thank
Nir Nice for supporting this work.

\bibliographystyle{abbrv}
\bibliography{paper}

\end{document}